\def\year{2017}\relax
\documentclass[letterpaper]{article}
\usepackage{aaai17}
\usepackage{times}
\usepackage{helvet}
\usepackage{courier}
\usepackage{url}
\usepackage{amsmath}
\usepackage{graphicx}
\usepackage{subcaption}
\usepackage{array}
\usepackage{multirow}
\usepackage{CJK}

\newcolumntype{L}[1]{>{\raggedright\arraybackslash}p{#1}}
\newcolumntype{C}[1]{>{\centering\arraybackslash}m{#1}}

\begin{document}
%
\title{A Dependency-Based Neural Reordering Model for \\Statistical Machine Translation}
\author{Christian Hadiwinoto \hspace{1cm} Hwee Tou Ng \\
Department of Computer Science \\
National University of Singapore \\
\{chrhad, nght\}@comp.nus.edu.sg\\
}
\maketitle
\begin{abstract}
In machine translation (MT) that involves translating between two languages with significant differences in word order, determining the correct word order of translated words is a major challenge. The dependency parse tree of a source sentence can help to determine the correct word order of the translated words.  In this paper, we present a novel reordering approach utilizing a neural network and dependency-based embeddings to predict whether the translations of two source words linked by a dependency relation should remain in the same order or should be swapped in the translated sentence. Experiments on Chinese-to-English translation show that our approach yields a statistically significant improvement of 0.57 BLEU point on benchmark NIST test sets, compared to our prior state-of-the-art statistical MT system that uses sparse dependency-based reordering features.
\end{abstract}

\section{Introduction}
In a machine translation (MT) system, determining the correct word order of translated words is crucial as word order reflects meaning. As different languages have different ordering of words, reordering of words is required to produce the correct translation output. Reordering in MT remains a major challenge for language pairs with a significant word order difference. Phrase-based MT systems \cite{koehn_statistical_2003}, which achieve state-of-the-art performance, generally adopt a reordering model based on the span of a phrase and the span of its adjacent phrase \cite{tillmann_unigram_2004,koehn_edinburgh_2005,galley_simple_2008}.

Incorporating the dependency parse tree of an input (source) sentence is beneficial for reordering, as the dependency tree captures the relationships between words in a sentence, through the dependency relation label between two words. The dependency parse tree of a source sentence can be utilized in reordering integrated within phrase-based statistical MT (SMT), by defining dependency-based features in the SMT log-linear model \cite{chang_discriminative_2009,hadiwinoto_swap_2016}.

Recently, neural networks have been applied to natural language processing (NLP) to minimize feature engineering and to utilize continuous word representation. It has found application in MT reordering, applied in the re-ranking of translation output candidates \cite{li_neural_2014,cui_lstm_2016} and in the pre-ordering approach (reordering the source sentence before translation) \cite{de_gispert_fast_2015,miceli-barone_non-projective_2015}. Nevertheless, reordering integrated with translation has not benefited from the neural approach.

In this paper, we propose applying a neural network (NN) model in reordering integrated within translation. We apply our neural classifier on two words linked by a dependency relation link, either a head-child or sibling link, in order to predict if two words need to be swapped in the translation. The prediction is used to guide the decoding process of state-of-the-art phrase-based SMT.

\section{A Neural Classifier for Dependency-Based Reordering}

We propose two neural classifiers, one to predict the correct order of the translated target words of two source words with a head-child relation, and the other for two source words with a sibling relation. Each binary classifier takes a set of features related to the two source words as its input and predicts if the translated words should be swapped (positive) or remain in order (negative).

\begin{figure*}[htb]
\centering
\begin{subfigure}[t]{0.2\textwidth}
\includegraphics[width=\textwidth]{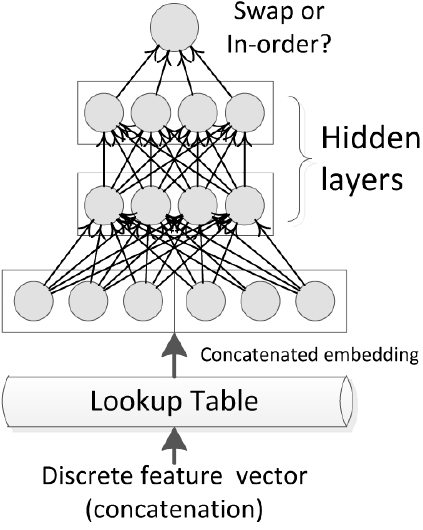}
\caption{~}
\label{fig:illustration_archi}
\end{subfigure}
\begin{subfigure}[t]{0.52\textwidth}
\includegraphics[width=\textwidth]{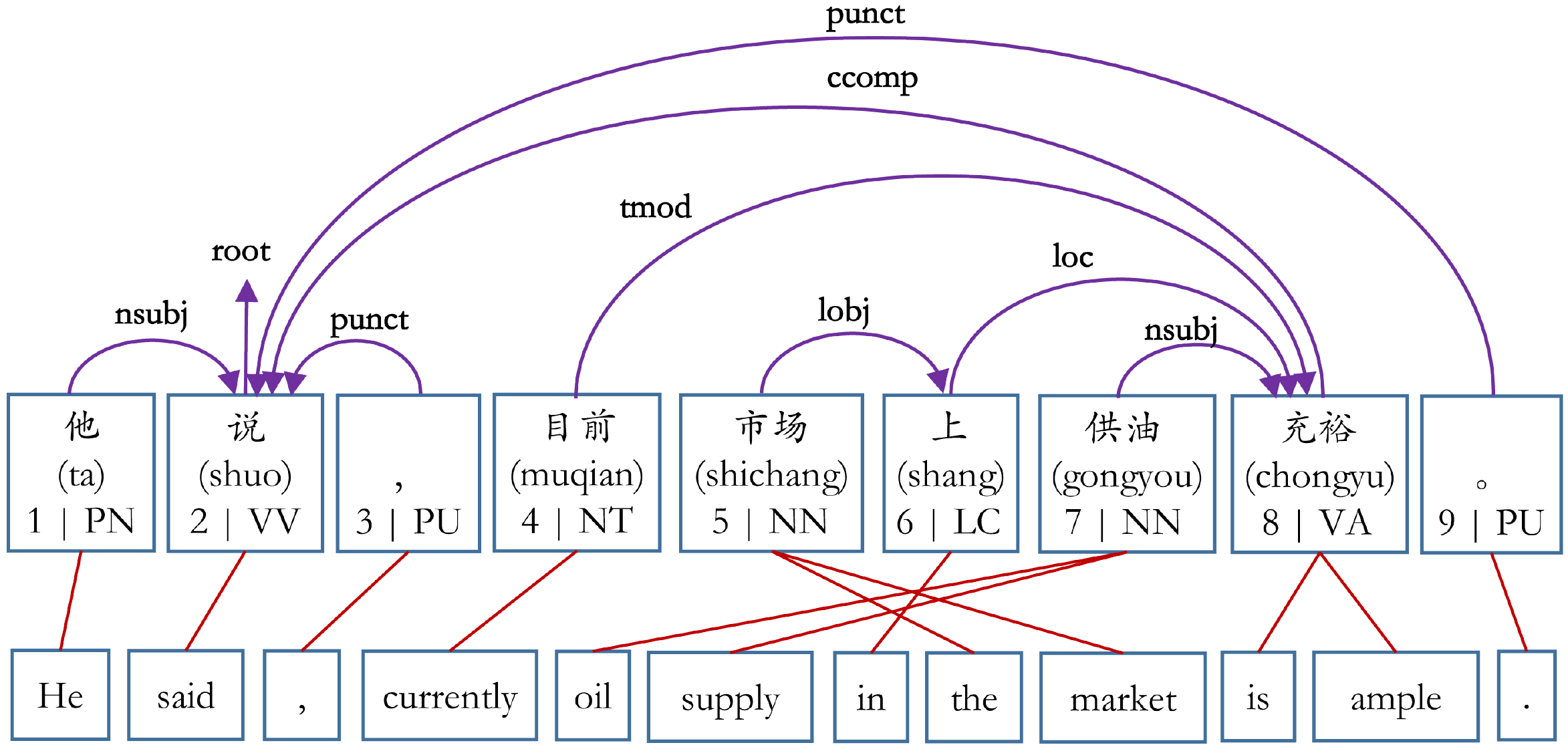}
\caption{~}
\label{fig:illustration_align}
\end{subfigure}
\begin{subfigure}[b]{1\textwidth}
{\centering
\footnotesize
\begin{tabular}{|c|l|l|l|l|l|l|l|l|l|r|r|c|}
\hline
\multicolumn{1}{|c|}{\multirow{2}{*}{Pair}} & \multicolumn{3}{c|}{Head} & \multicolumn{3}{c|}{Child, if left} & \multicolumn{3}{c|}{Child, if right} & \multicolumn{1}{c|}{Dist.} & \multicolumn{1}{c|}{Punct.} & \multicolumn{1}{c|}{\multirow{2}{*}{Label}} \\
\cline{2-12}
 & \multicolumn{1}{c|}{$x_h$} & \multicolumn{1}{c|}{$T(x_h)$} & \multicolumn{1}{c|}{$L(x_h)$} & \multicolumn{1}{c|}{$x_c$} & \multicolumn{1}{c|}{$T(x_c)$} & \multicolumn{1}{c|}{$L(x_c)$} & \multicolumn{1}{c|}{$x_c$} & \multicolumn{1}{c|}{$T(x_c)$} & \multicolumn{1}{c|}{$L(x_c)$} & \multicolumn{1}{c|}{$d(x_h,x_c)$} & \multicolumn{1}{c|}{$\omega(x_h,x_c)$} &  \\
\hline
(2, 1) & shuo & VV & root & ta & PN & nsubj & NULL & NULL & NULL & $-1$ & $0$ & $0$\\
(2, 8) & shuo & VV & root & NULL & NULL & NULL & chongyu & VA & ccomp & $+2$ & $1$ & $0$\\
(6, 5) & shang & LC & loc & shichang & NN & lobj & NULL & NULL & NULL & $-1$ & $0$ & $1$\\
\hline
\end{tabular}
\caption{~}
\label{fig:illustration_hcfeat}}
\end{subfigure}
\begin{subfigure}[b]{1\textwidth}
{\centering
\footnotesize
\begin{tabular}{|c|l|l|l|r|l|l|l|r|l|l|r|c|}
\hline
\multicolumn{1}{|c|}{\multirow{2}{*}{Pair}} & \multicolumn{4}{c|}{Left child} & \multicolumn{4}{c|}{Right child} & \multicolumn{2}{c|}{Head} & \multicolumn{1}{c|}{Punct.} & \multicolumn{1}{c|}{\multirow{2}{*}{Label}} \\
\cline{2-12}
 & \multicolumn{1}{c|}{$x_l$} & \multicolumn{1}{c|}{$T(x_l)$} & \multicolumn{1}{c|}{$L(x_l)$} & \multicolumn{1}{c|}{$d(x_h,x_l)$} & \multicolumn{1}{c|}{$x_r$} & \multicolumn{1}{c|}{$T(x_r)$} & \multicolumn{1}{c|}{$L(x_r)$} & \multicolumn{1}{c|}{$d(x_h,x_r)$} & \multicolumn{1}{c|}{$x_h$} & \multicolumn{1}{c|}{$T(x_h)$} & \multicolumn{1}{c|}{$\omega(x_l,x_r)$} & \\
\hline
(1,8) & ta & PN & nsubj & $-1$ & chongyu & VA & ccomp & $+2$ & shuo & VV & $1$ & $0$ \\
(4,6) & muqian & NT & tmod & $-2$ & shang & LC & loc & $-2$ & chongyu & VA & $0$ & $0$ \\
(6,7) & shang & LC & loc & $-2$ & gongyou & NN & nsubj & $-1$ & chongyu & VA & $0$ & $1$ \\
\hline
\end{tabular}
\caption{~}
\label{fig:illustration_sibfeat}}
\end{subfigure}
\caption{\label{fig:illustration} Illustration: (a) neural network classifier architecture with two hidden layers; (b) an aligned Chinese-English parallel sentence pair; and sample extracted training instances and features for (c) head-child classifier and (d) sibling classifier. The label 1 or 0 indicates whether the two words need to be swapped or kept in order, respectively.}
\end{figure*}

\subsection{Input Representation}
The head-child classifier predicts the order of the translated words of a source word $x_c$ and its head word $x_h$ (where $x_g$ is the head word of $x_h$) using the following input features:
\begin{itemize}
\item The head word $x_h$, its part-of-speech (POS) tag $T(x_h)$, and the dependency label $L(x_h)$ linking $x_h$ to $x_g$
\item The child word $x_c$, its POS tag $T(x_c)$, and the dependency label $L(x_c)$ linking $x_c$ to $x_h$
\item The signed distance $d(x_h,x_c)$ between the head and the child in the original source sentence, with the following possible values:
\begin{itemize}
\item $-2$ if $x_c$ is on the left of $x_h$ and there is at least one other child between them
\item $-1$ if $x_c$ is on the left of $x_h$ and there is no other child between them
\item $+1$ if $x_c$ is on the right of $x_h$ and there is no other child between them
\item $+2$ if $x_c$ is on the right of $x_h$ and there is at least one other child between them
\end{itemize}
\item A Boolean $\omega(x_h,x_c)$ to indicate if any punctuation symbol, which is also the child of $x_h$, exists between $x_h$ and $x_c$
\end{itemize}

The sibling classifier predicts the order of the translated words of two source words $x_l$ and $x_r$, where $x_l$ is to the left of $x_r$ and both have the common head word $x_h$, using the following features:
\begin{itemize}
\item The left child word $x_l$, its POS tag $T(x_l)$, the dependency label $L(x_l)$ linking $x_l$ to $x_h$, and the signed distance to its head $d(x_h,x_l)$ 
\item The right child word $x_r$, its POS tag $T(x_r)$, the dependency label $L(x_r)$ linking $x_r$ to $x_h$, and the signed distance to its head $d(x_h,x_r)$ 
\item The head word $x_h$ and its POS tag $T(x_h)$
\item A Boolean $\omega(x_l,x_r)$ to indicate if any punctuation symbol, which is also the child of $x_h$, exists between $x_l$ and $x_r$
\end{itemize}

\subsection{Feed-Forward Layers}

As shown in Figure \ref{fig:illustration_archi}, the classifier is a feed-forward neural network whose input layer contains the features. Each feature is mapped by a lookup table to a continuous vector representation, and the resulting vectors are concatenated and fed into (multiplied by) a series of hidden layers (weight matrices) using the rectified linear activation function, $relu(x)=\max(0,x)$. Given the hidden-layer-transformed embedding vector ${\bf x}$, a weight vector ${\bf W}$, and a bias value $b$, the prediction output $\sigma$ is defined as:
\begin{eqnarray}
z &=& {\bf W\cdot{}x}+b \\
\sigma(z) &=& \frac{1}{1+e^{-z}}
\end{eqnarray}
We initialize the hidden layers and the embedding layer for non-word features (POS tags, dependency labels, and Boolean indicators) by a random uniform distribution. For word features $x_h$, $x_c$, $x_l$, and $x_r$, we initialize their embeddings by the dependency-driven embedding scheme of \cite{bansal_tailoring_2014}. This scheme is a modified skip-gram model, which given an input word, predicts its context (surrounding words), resulting in a mapping such that words with similar surrounding words have similar continuous vector representations \cite{mikolov_efficient_2013}. Similarly, defining the dependency information (i.e., label, head word, and child word) as context produces a mapping such that words with similar head and child words have similar continuous vector representations.

Dependency-driven embedding can be obtained from a dependency-parsed corpus, where each training instance is formulated as (note: $x_g$ is the head of $x_h$):
\[
L(x_{h})_{<GL>}\hspace{0.1cm}{{x}_{g}}_{<G>}\hspace{0.1cm}x_h\hspace{0.1cm}x_c\hspace{0.1cm}L(x_c)_{<L>}
\]
The skip-gram model is trained with a window size of 1 (denoting one context item on the left and one on the right). Following \cite{bansal_tailoring_2014}, the items marked by $<>$ subscripts serve as the context, have different continuous vector representations from the words ($x_h$ and $x_c$), and are filtered out from the embedding vocabulary after training.

\subsection{Neural Network Training}

The training instances for the neural classifiers are obtained from a word-aligned parallel corpus. Two source-side words with head-child or sibling relation are extracted with their corresponding order label, \textit{swapped} or \textit{in order}, depending on the positions of their aligned target-side words. Figure \ref{fig:illustration} shows the training instances extracted with their corresponding features. For the head-child classifier, the features containing the child information are distinguished based on whether the child is on the left or right of the head.

The NN classifiers are trained using back-propagation to minimize the cross-entropy objective function:
\begin{equation}
L=-\frac{1}{T}\sum_{i=1}^{T} y_i \log\hat{y_i} + (1-y_i) \log(1-\hat{y_i})
\end{equation}
where $x_i$ is the $i$-th training instance, $y_i$ is its corresponding label ($1$ for \textit{swapped} and $0$ for \textit{in order}), and $\hat{y_i}$ is the classifier prediction probability for \textit{swapped}. To prevent model overfitting, we used the dropout strategy \cite{srivastava_dropout:_2014} on the input embedding layer.

\section{Reordering in Phrase-Based SMT}

We adopt the phrase-based SMT approach, using a beam search decoding algorithm \cite{koehn_pharaoh:_2004}. Each source phrase and one of its possible translations represent an alternative in the search for the translated sentence. While the search produces translation from left to right in the translation output order, it picks the source phrases in any order to enable reordering for a language pair with different word order. The translation output is picked based on a score computed by a log-linear model, comprising the weighted sum of feature function values \cite{och_discriminative_2002}.

A phrase-based SMT system typically includes a distance-based reordering penalty (DBR) \cite{koehn_statistical_2003}, to discourage long-distance reordering, and phrase-based reordering models (PBRM). The latter comprises the phrase-based lexicalized reordering (PBLR) model \cite{tillmann_unigram_2004,koehn_edinburgh_2005} and the hierarchical reordering (HR) model \cite{galley_simple_2008}. These models are the conventional reordering models widely used in phrase-based SMT.

\subsection{Dependency-Based Decoding Features}

Phrase-based decoding can take into account the source dependency parse tree to guide its search. To encourage structural cohesion during translation, we add a dependency distortion penalty (DDP) feature \cite{cherry_cohesive_2008} to discourage translation output in which the translated words of a phrase in a source dependency parse subtree are split.

We also incorporate the sparse dependency swap (DS) features of our prior work \cite{hadiwinoto_swap_2016}. The features involve considering a source word $x$ being translated during beam search and each of the as yet untranslated source words $x'$, where $x'$ is the head, the child, or the sibling of $x$ in the source dependency parse tree. $x'$ can be on the left of $x$ in the source sentence, resulting in $x$ and $x'$ being \textit{swapped} in the translation output; or $x'$ can be on the right of $x$, resulting in $x$ and $x'$ following the same order (\textit{in order}). This principle is used to guide word pair translation ordering through sparse feature templates for head-child word pair and sibling word pair, in which each word $x$ in a word pair is represented by its dependency label ($L(x)$), POS tag ($T(x)$), and their combination.

Specifically, the sparse dependency swap (DS) features of \cite{hadiwinoto_swap_2016} are based on a feature template for a head word $x_h$ and its child word $x_c$, their dependency labels and POS tags, whether $x_h$ is on the $p\nolinebreak\in\nolinebreak\{left,right\}$ of $x_c$ in the source sentence, and the ordering of the pair in the translation output $o\nolinebreak\in\nolinebreak\{in\_order,swapped\}$:
\begin{equation}
H_{hc}(x_h,x_c,p,o) =
\left[\begin{array}{c}
h_{hc}(L(x_h),L(x_c),p,o) \\
h_{hc}(T(x_h),T(x_c),p,o) \\
h_{hc}(L(x_h),T(x_c),p,o) \\
h_{hc}(T(x_h),L(x_c),p,o)
\end{array}\right]
\label{eq:temp_parchild}
\end{equation}
Similarly, there is another feature template for two sibling words, $x_l$ on the left of $x_r$ sharing a common head word, their dependency labels and POS tags, and the ordering of the pair in the translation output $o\nolinebreak\in\nolinebreak\{in\_order,swapped\}$:
\begin{equation}
H_{sib}(x_l,x_r,o) =
\left[\begin{array}{c}
h_{sib}(L(x_l),L(x_r),o) \\
h_{sib}(T(x_l),T(x_r),o) \\
h_{sib}(L(x_l),T(x_r),o) \\
h_{sib}(T(x_l),L(x_r),o)
\end{array}\right]
\label{eq:temp_sibling}
\end{equation}

\subsection{Incorporating Neural Classifier}

We incorporate the neural classifier by defining one decoding feature function for the head-child classifier, and another decoding feature function for the sibling classifier. We also employ model ensemble by training multiple head-child and sibling classifiers, each with a different random seed for hidden layer initialization. Within the log-linear model, the value of each neural classifier feature function is its prediction log-probability. Each feature function is assigned a different weight obtained from tuning on development data.

\section{Experimental Setup}
\subsection{Data Set and Toolkits}

We conducted experiments on a phrase-based Chinese-to-English SMT system built using Moses \cite{koehn_moses:_2007}. Our parallel training corpora are from LDC, which we divide into older corpora\footnote{LDC2002E18, LDC2003E14, LDC2004E12, LDC2004T08, LDC2005T06, and LDC2005T10.} and newer corpora\footnote{LDC2007T23, LDC2008T06, LDC2008T08, LDC2008T18, LDC2009T02, LDC2009T06, LDC2009T15, LDC2010T03, LDC2013T11, LDC2013T16, LDC2014T04, LDC2014T11, LDC2014T15, LDC2014T20, and LDC2014T26.}. Due to the dominant older corpora, we duplicate the newer corpora of various domains ten times to achieve better domain balance. To reduce the possibility of alignment errors, parallel sentences in the corpora that are longer than 85 words in either Chinese (after word segmentation) or English are discarded. In the end, the final parallel texts consist of about 8.8M sentence pairs, 228M Chinese tokens, and 254M English tokens (a token can be a word or punctuation symbol). We also added two dictionaries\footnote{LDC2002L27 and LDC2005T34.}, having 1.81M Chinese tokens and 2.03M English tokens in total, by concatenating them to our training parallel texts. To train the Chinese word embeddings as described above, we concatenate the Chinese side of our parallel texts with Chinese Gigaword version 5 (LDC2011T13), resulting in 2.08B words in total.

All Chinese sentences in our experiment are first word-segmented using a maximum entropy-based Chinese word segmenter \cite{low_maximum_2005} trained on the Chinese Treebank (CTB) segmentation standard. Then the parallel corpus is word-aligned by GIZA++ \cite{och_systematic_2003} using IBM Models 1, 3, and 4 \cite{brown_mathematics_1993}\footnote{The default when running GIZA++ with Moses.}. For building the phrase table, which follows word alignment, the maximum length of a phrase is set to 7 words for both the source and target sides.

The language model (LM) is a 5-gram model trained on the English side of the FBIS parallel corpus (LDC2003E14) and the monolingual corpus English Gigaword version 4 (LDC2009T13), consisting of 107M sentences and 3.8B tokens altogether. Each individual Gigaword sub-corpus\footnote{AFP, APW, CNA, LTW, NYT, and Xinhua} is used to train a separate language model and so is the English side of FBIS. These individual language models are then interpolated to build one single large LM, via perplexity tuning on the development set.

Training the neural reordering classifier involves LDC manually-aligned corpora, from which we extracted 572K head-child pairs and 1M sibling pairs as training instances\footnote{LDC2012T20, LDC2012T24, LDC2013T05, LDC2013T23, LDC2014T25, LDC2015T04, and LDC2015T18.}, while retaining 90,233 head-child pairs and 146,112 sibling pairs as held-out tuning instances\footnote{LDC2012T16.}. The latter is used to pick the best neural network parameters.

Our translation development set is MTC corpus version 1 (LDC2002T01) and version 3 (LDC2004T07). This development set has 1,928 sentence pairs in total, 49K Chinese tokens and 58K English tokens on average across the four reference translations. Weight tuning is done by using the pairwise ranked optimization (PRO) algorithm \cite{hopkins_tuning_2011}.

We parse the Chinese sentences by the Mate parser, which jointly performs POS tagging and dependency parsing \cite{bohnet_transition-based_2012}, trained on Chinese Treebank (CTB) version 8.0 (LDC2013T21).

Our translation test set consists of the NIST MT evaluation sets from 2002 to 2006, and 2008\footnote{LDC2010T10, LDC2010T11, LDC2010T12, LDC2010T14, LDC2010T17, and LDC2010T21.}.

\subsection{Baseline System}

Our phrase-based baseline SMT system includes the conventional reordering models, i.e., distance-based reordering penalty (DBR) and phrase-based reordering model (PBRM), both phrase-based lexicalized reordering (PBLR) and hierarchical reordering (HR). We also use the dependency-based reordering features, including the distortion penalty (DDP) feature and the sparse dependency swap (DS) features.

To constrain the decoding process, we set punctuation symbols as reordering constraint across which phrases cannot be reordered, as they form the natural boundaries between different clauses. In addition, a distortion limit is set such that reordering cannot be longer than a certain distance. To pick the translation output, we also use $n$-best minimum Bayes risk (MBR) decoding \cite{kumar_minimum_2004} instead of the default maximum a-posteriori (MAP) decoding.

\subsection{Neural Reordering}

We replaced DS features by our dependency-based neural reordering classifier, in which we set the word vocabulary to the 100,000 most frequent words in our parallel training corpora, replacing other words with a special $UNK$ token, in addition to all POS tags, dependency labels, and Boolean features. We set the embedding dimension size to 100, the lower hidden layer dimension size to 200, and the upper hidden layer dimension size to 100. We trained for 100 epochs, with 128 mini-batches per epoch, and used a dropout rate of $0.5$. For model ensemble, we trained 10 classifiers for head-child reordering and 10 for sibling reordering, each of which forming one feature function.

\begin{table*}[ht]
\centering
\small
\begin{tabular}{|l|l|l||l|l||l|l|l|l|l|}
\hline
\multicolumn{1}{|c|}{\multirow{4}{*}{\bf Dataset}} & \multicolumn{4}{c||}{\bf Baseline} & \multicolumn{5}{c|}{\multirow{2}{*}{\bf Neural reordering classifier}} \\\cline{2-5}
 & \multicolumn{2}{c||}{\bf Conventional} & \multicolumn{2}{c||}{\bf Sparse dependency} &  \multicolumn{5}{c|}{} \\\cline{2-10}
 & \multicolumn{1}{c|}{\multirow{2}{*}{\bf DBR}} & \multicolumn{1}{c||}{\bf DBR} & \multicolumn{1}{c|}{\multirow{2}{*}{\bf DDP+DS}} & \multicolumn{1}{c||}{\bf DBR+PBRM} & \multicolumn{1}{c|}{\bf DDP+NR} & \multicolumn{4}{c|}{\bf DBR+PBRM+DDP+NR} \\\cline{6-10}
 & & \multicolumn{1}{c||}{\bf +PBRM} & & \multicolumn{1}{c||}{\bf +DDP+DS} & \multicolumn{1}{c|}{\bf 10s-D} & \multicolumn{1}{c|}{\bf 1s-D} & \multicolumn{1}{c|}{\bf 10s-R} & \multicolumn{1}{c|}{\bf 10s-S} & \multicolumn{1}{c|}{\bf 10s-D}\\
\hline\hline
{\bf Dev} & 37.44 & 40.04 & 40.80 & 41.39 & 40.76 & 40.96 & 40.18 & 40.80 & 41.21\\\hline
\hline
{\bf NIST02} & 37.23 & 39.19 & 39.96$^{**\dagger}$ & 40.48$^{**\dagger\dagger}$ & 40.85$^{**\dagger\dagger}$ & 40.59$^{**\dagger\dagger}$ & 40.07$^{**\dagger\dagger}$ & 40.57$^{**\dagger\dagger}$ & 40.87$^{**\dagger\dagger\ddagger}$\\\hline
{\bf NIST03} & 37.24 & 39.44 & 39.72$^{**}$ & 40.88$^{**\dagger\dagger}$ & 40.75$^{**\dagger\dagger}$ & 41.11$^{**\dagger\dagger}$ & 40.15$^{**\dagger\dagger}$ & 40.67$^{**\dagger\dagger}$ & 41.46$^{**\dagger\dagger\ddagger\ddagger}$\\\hline
{\bf NIST04} & 37.66 & 40.26 & 40.04$^{**}$ & 40.97$^{**\dagger\dagger}$ & 40.85$^{**\dagger\dagger}$ & 41.29$^{**\dagger\dagger\ddagger}$ & 40.65$^{**\dagger}$ & 41.00$^{**\dagger\dagger}$ & 41.70$^{**\dagger\dagger\ddagger\ddagger}$\\\hline
{\bf NIST05} & 37.32 & 39.65 & 39.34$^{**}$ & 41.26$^{**\dagger\dagger}$ & 40.79$^{**\dagger\dagger}$ & 41.32$^{**\dagger\dagger}$ & 40.12$^{**}$ & 41.37$^{**\dagger\dagger}$ & 41.56$^{**\dagger\dagger}$\\\hline
{\bf NIST06} & 36.15 & 38.70 & 37.79$^{**}$ & 39.15$^{**\dagger}$ & 39.25$^{**}$ & 39.44$^{**\dagger\dagger\ddagger}$ & 38.00$^{**}$ & 39.46$^{**\dagger\dagger}$ & 39.95$^{**\dagger\dagger\ddagger\ddagger}$\\\hline
{\bf NIST08} & 28.47 & 30.11 & 29.57$^{**}$ & 31.12$^{**\dagger\dagger}$ & 31.17$^{**\dagger\dagger}$ & 32.02$^{**\dagger\dagger\ddagger\ddagger}$ & 29.70$^{**}$ & 31.28$^{**\dagger\dagger}$ & 31.76$^{**\dagger\dagger\ddagger\ddagger}$\\\hline
\hline
{\bf Average} & 35.68 & 37.89 & 37.74$^{**}$ & 38.98$^{**\dagger\dagger}$ & 38.94$^{**\dagger\dagger}$ & 39.30$^{**\dagger\dagger}$ & 38.12$^{**\dagger\dagger}$ & 39.06$^{**\dagger\dagger}$ & {\bf 39.55}$^{**\dagger\dagger\ddagger\ddagger}$\\\hline
\end{tabular}
\caption{\label{tab:mainresult} BLEU scores (\%) of our neural reordering ({\bf NR}) approach, either single system ({\bf 1s}) or 10-system ensemble ({\bf 10s}), with different embedding initialization schemes, i.e., random ({\bf R}), skip-gram ({\bf S}), or dependency-driven ({\bf D}). We used the following prior reordering features as baseline: (1) distance-based reordering ({\bf DBR}) \cite{koehn_statistical_2003}; (2) phrase-based reordering models ({\bf PBRM}), comprising phrase-based lexicalized reordering \cite{tillmann_unigram_2004,koehn_edinburgh_2005} and hierarchical reordering \cite{galley_simple_2008}; (3) dependency distortion penalty ({\bf DDP}) \cite{cherry_cohesive_2008}; and (4) sparse dependency swap features ({\bf DS}) \cite{hadiwinoto_swap_2016}. Statistical significance testing compares our approach with {\bf DBR} ($*$: significant at $p < 0.05$; $**$: significant at $p < 0.01$), with {\bf DBR+PBRM} ($\dagger$: significant at $p < 0.05$; $\dagger\dagger$: significant at $p < 0.01$), and with {\bf DBR+PBRM+DDP+DS} ($\ddagger$: significant at $p < 0.05$; $\ddagger\ddagger$: significant at $p < 0.01$).}
\end{table*}

\section{Experimental Results}

The translation quality of the system output is measured by case-insensitive BLEU \cite{papineni_bleu:_2002}, for which the brevity penalty is computed based on the shortest reference (NIST-BLEU)\footnote{\url{ftp://jaguar.ncsl.nist.gov/mt/resources/mteval-v11b.pl}}. Statistical significance testing between systems is conducted by bootstrap resampling \cite{koehn_statistical_2004}.

Table \ref{tab:mainresult} shows the experimental results. The distortion limit of all the systems is set to 14. As shown in the table, when the word embedding features are initialized using dependency context \cite{bansal_tailoring_2014}, which is our default scheme, our translation system with single neural classifier is able to improve over our strong baseline system (DBR+PBRM+DDP+DS) by +0.32 BLEU point, while an ensemble model of 10 neural classifiers improves over our baseline system by +0.57 BLEU point. The results show that the neural reordering classifier is able to replace the sparse dependency swap features and achieves better performance.

In addition to the dependency-driven embedding initialization scheme of \cite{bansal_tailoring_2014}, we are also interested in testing other word embedding schemes. Additional experiments use two other initialization schemes: (1) random initialization and (2) the original skip-gram model of \cite{mikolov_efficient_2013} with a window size of 5. As shown in Table \ref{tab:mainresult}, using dependency-driven embedding initialization scheme yields the best improvement over our baseline. On the other hand, random initialization of word embedding yields worse results compared to our baseline, showing a significant drop. Using the skip-gram word embedding model yields average results comparable to the baseline.

We are also interested in testing the performance of the dependency-based reordering features in the absence of the conventional phrase-based reordering models. Table \ref{tab:mainresult} shows that the system with dependency distortion penalty (DDP) and sparse dependency swap (DS) features is unable to outperform the system with only conventional phrase-based reordering models (DBR+PBRM). However, our neural classifier approach, without the conventional reordering models, significantly outperforms the conventional reordering models by +1.05 BLEU point.

\section{Discussion}

Dependency swap features capture the dependency label and POS tag of the two words to be reordered, but not the actual words themselves. While using words as sparse features may result in too many parameters, the continuous word representation in our neural approach alleviates this problem. In addition, the neural network model also learns useful combinations of individual features. While dependency swap features \cite{hadiwinoto_swap_2016} define features as pairs of dependency label and POS tag, the hidden layer of a NN can dynamically choose the information to take into account for the reordering decision.

Using neural classifiers with dependency-based word embedding initialization yields significant improvement, whereas random initialization and skip-gram initialization of word embeddings yield no improvement. This shows the importance of capturing dependency information in the word embeddings for reordering.

\begin{CJK}{UTF8}{gbsn}
\begin{figure*}
\begin{subfigure}[b]{0.36\textwidth}
\footnotesize
\begin{tabular}{|L{\textwidth}|}
\hline
{\bf \em Source sentence} \newline
ISO\hspace{0.1cm}是\hspace{0.1cm}目前\hspace{0.1cm}世界\hspace{0.1cm}上\hspace{0.1cm}两\hspace{0.1cm}大\hspace{0.1cm}国际\hspace{0.1cm}标准化\hspace{0.1cm}组织\hspace{0.1cm}之一\hspace{0.1cm}。\\
\hline
{\bf \em Reference translation:} \newline
ISO is one of the world's two international standardization organizations.\\
\hline
{\bf \em DBR+PBRM:}\newline
The two major International Organization for Standardization (ISO is one of the world.\\
\hline

{\bf \em DBR+PBRM+DDP+DS:}\newline
The International Organization for Standardization (ISO is one of the two in the world.\\
\hline

{\bf \em DDP+NR, DBR+PBRM+DDP+NR:}\newline
The ISO is one of the two major international standardization organization in the world. \\
\hline
\end{tabular}

\end{subfigure}
\hfill
\begin{subfigure}[c]{0.6\textwidth}
{\centering
\small
\includegraphics[width=\textwidth]{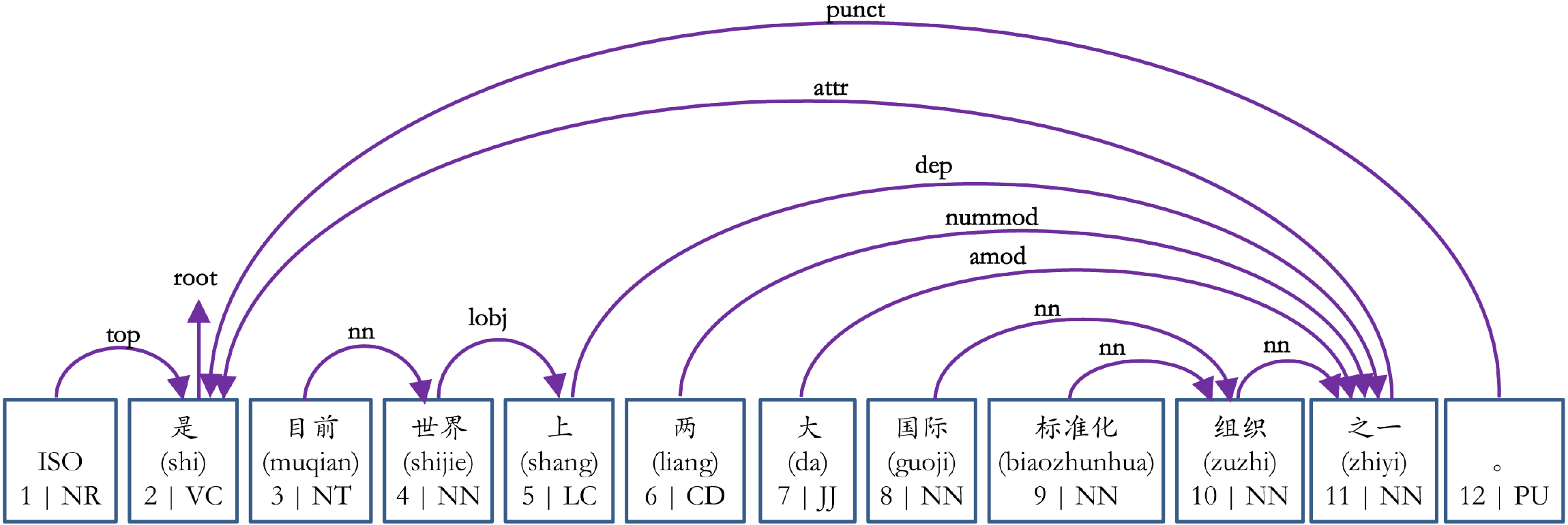}}
\end{subfigure}
\caption{\label{fig:comparison} {\bf Left}: a sample sentence and our translation output with distance-based reordering ({\bf DBR}), phrase-based reordering models ({\bf PBRM}), dependency distortion penalty ({\bf DDP}), and sparse dependency swap ({\bf DS}) features, compared to neural reordering in a 10-system ensemble with dependency-driven embeddings ({\bf NR}); {\bf Right}: a source dependency parse tree.}
\end{figure*}
\end{CJK}

Figure \ref{fig:comparison} shows the baseline phrase-based SMT system with conventional phrase-based reordering models (DBR+PBRM) and sparse dependency swap features produces an incorrect translation output. The sparse dependency swap features prefer the Chinese words ``zhiyi (one of)'' and ``zuzhi (organization)'', where ``zhiyi'' is the head word of ``zuzhi'', to remain in order after translation, based on their dependency labels and POS tags. However, the Chinese expression ``$NOUN$ zhiyi'' should be swapped in the English translation, resulting in ``one of $NOUN$''\footnote{The ordering is further aggravated by wrongly swapping ``ISO'' and ``zuzhi (organization)'', due to the translation output score being the weighted sum of features including LM, which prefers such a translation.}.

Our experimental results also show that without conventional phrase-based reordering models, the sparse dependency-based features are unable to outperform the conventional reordering models, whereas the neural dependency-based reordering model outperforms the conventional reordering models. This further demonstrates the strength of our dependency-based neural reordering approach.

Our approach applies syntax to SMT with beam search decoding. This is different from prior approaches requiring chart parsing decoding such as the hierarchical phrase-based \cite{chiang_hierarchical_2007}, tree-to-string \cite{liu_tree--string_2006}, string-to-tree \cite{marcu_spmt:_2006}, and tree-to-tree \cite{zhai_simple_2011} SMT approaches.

The end-to-end neural MT (NMT) approach has recently been proposed for MT. However, the most recent NMT papers tested on the same NIST Chinese-to-English test sets \cite{wang_memory-enhanced_2016,zhang_variational_2016} show lower absolute BLEU scores (by 2 to 7 points) compared to our scores. Following the approach of \cite{junczys-dowmunt_is_2016}, our own implemented NMT system (single system without ensemble), when trained on the same corpora and tested on the same NIST test sets in this paper, achieves an average BLEU score of 38.97, lower by 0.58 point compared to our best SMT system ($p<0.01$). This shows that our neural dependency-based reordering model outperforms the NMT approach. NMT also requires longer time to train (18 days) compared to our best SMT system (3 days).

\section{Related Work}

Phrase-based SMT reordering can utilize the dependency parse of the input sentence. \citeauthor{chang_discriminative_2009} \shortcite{chang_discriminative_2009} utilized the traversed paths of dependency labels to guide phrase reordering. \citeauthor{hadiwinoto_swap_2016} \shortcite{hadiwinoto_swap_2016} introduced a technique to determine the order of two translated words with corresponding source words that are related through the dependency parse during beam search. They defined sparse decoding features to encourage or penalize the reordering of two words, based on the POS tag and dependency relation label of each word, but not the words themselves.

Neural reordering models have been applied to re-rank translation candidates generated by the translation decoder. \citeauthor{li_neural_2014} \shortcite{li_neural_2014} introduced a recursive auto-encoder model to represent phrases and determine the phrase orientation probability. \citeauthor{cui_lstm_2016} \shortcite{cui_lstm_2016} introduced long short-term memory (LSTM) recurrent neural networks to predict the translation word orientation probability. These approaches did not use dependency parse and they were not applied directly during decoding.

Source dependency parse is also used in the pre-ordering approach, which pre-orders words in a source sentence into target word order and then translates the target-ordered source sentence into the target language. While the pre-ordering step typically utilizes a classifier with feature combinations \cite{lerner_source-side_2013,jehl_source-side_2014}, a neural network can replace the classifier to avoid feature combination. De Gispert, Iglesias, and Byrne \shortcite{de_gispert_fast_2015} introduced a feed-forward neural network to pre-order the dependency parse tree nodes (words). However, they did not explore dependency-driven embeddings and model ensemble. \citeauthor{miceli-barone_non-projective_2015} \shortcite{miceli-barone_non-projective_2015} treat pre-ordering as a traversal on the dependency parse tree, guided by a recurrent neural network. In these approaches, the translation possibility is limited to one target ordering. In contrast, applying a neural reordering model jointly with beam search allows for multiple ordering alternatives and interaction with other models, such as the phrase-based reordering models. We can even build multiple neural models (ensemble) and assign a different weight to each of them to optimize translation quality.

Our neural reordering classifier serves as a decoding feature function in SMT, leveraging the decoding. This is similar to prior work on neural decoding features, i.e., neural language model \cite{vaswani_decoding_2013} and neural joint model \cite{devlin_fast_2014}, a source-augmented language model. However, these features are not about word reordering.

While continuous representation of words is originally defined for words \cite{mikolov_efficient_2013}, we also define continuous representation for POS tags, dependency labels, and indicator features. Extending continuous representation to non-word features is also done in neural dependency parsing \cite{chen_fast_2014,andor_globally_2016}, which shows better performance by using continuous feature representation over the traditional discrete representation.

\section{Conclusion}

We have presented a dependency-based reordering approach for phrase-based SMT, guided by neural classifier predictions. It shows that MT can be improved by a neural network approach by not requiring explicit feature combination and by using dependency-driven continuous word representation. Our experiments also show that our neural reordering approach outperforms our prior reordering approach employing sparse dependency-based features.

\bibliographystyle{aaai}
\bibliography{nnr}

\end{document}